\documentclass[11pt]{article}

\usepackage[preprint]{acl}

\usepackage{times}
\usepackage{latexsym}

\usepackage{kotex}
\usepackage{MnSymbol}
\usepackage{graphicx}
\usepackage{booktabs}
\usepackage{makecell}
\usepackage[most]{tcolorbox}
\usepackage{xcolor}

\usepackage[T1]{fontenc}
\usepackage[utf8]{inputenc}

\usepackage[verbose=silent]{microtype}

\usepackage{inconsolata}

\usepackage{graphicx}

\title{Decompose-and-Refine: Structured Legal Question Answering with Parametric Retrieval}

\author{
    Jihyung Lee
    $^1$, 
    \textbf{Hyounghun Kim}$^{1,2}$, 
    \textbf{Gary Geunbae Lee}$^{1,2}$ \\
    $^1$Graduate School of Artificial Intelligence, POSTECH, Republic of Korea\\
    $^2$Department of Computer Science and Engineering, POSTECH, Republic of Korea\\
    \texttt{\{jihyung.lee, 
    h.kim, 
    gblee\}@postech.ac.kr} \\
}

\begin{document}
\maketitle

\newcommand{\methodName}{\textsc{DaR}}

\begin{abstract}
Large language models (LLMs) have shown strong performance in the legal domain, demonstrating notable potential in Legal Question Answering (LQA). However, unlike general QA, LQA requires answers that are not only accurate but also rigorously grounded in explicit legal authority. In statutory LQA, many questions require multi-hop reasoning across multiple legal issues, substantially increasing the risk of hallucination, thereby making accurate retrieval of supporting statutory provisions a critical prerequisite. Despite recent progress in multi-hop QA, existing approaches often rely on reasoning in natural language or retrieval without explicit query reformulation, leaving the vocabulary gap between user questions and statutory text largely unaddressed. 
To address this challenge, we propose \textbf{Decompose-and-Refine (\methodName{})}, a statute-grounded LQA framework that tightly integrates step-wise question decomposition with parametric knowledge–based query refinement. \methodName{} progressively decomposes a complex legal question into atomic sub-questions and generates statute-aligned parametric queries for each sub-question, enabling the selection of a single most central statutory provision corresponding to each legal issue.
We evaluate \methodName{} on KoBLEX, a Korean multi-hop LQA benchmark grounded in statutory law, using Qwen3-32B and Gemma3-27B. Experimental results demonstrate that \methodName{} consistently improves both retrieval accuracy and final answer quality over existing approaches. Moreover, by explicitly separating sub-questions and their corresponding statutory provisions, \methodName{} facilitates transparent, issue-level verification of complex legal reasoning processes.
\end{abstract}

\section{Introduction}
\begin{figure}[t]
    \centering
    \includegraphics[width=\linewidth]{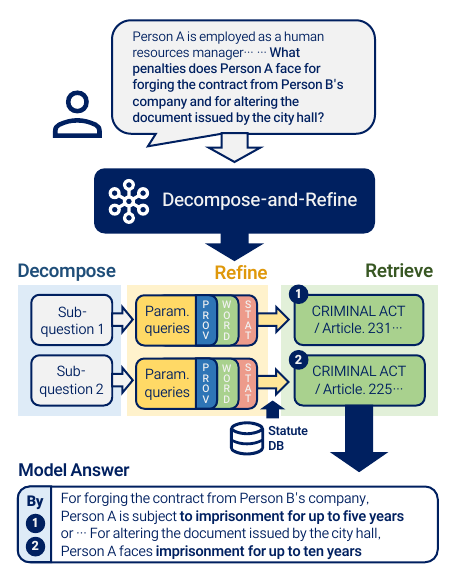}
    \caption{Overview of Decompose-and-Refine (\methodName{}).
\methodName{} decomposes a complex legal question into atomic sub-questions, generates three types of parametric queries for each sub-question, and retrieves exactly one core statutory provision per issue, enabling controlled and evidence-centric legal reasoning.} 
    \label{fig:overview}
\end{figure}

Large language models (LLMs) have demonstrated strong performance across a wide range of tasks in the legal domain,
showing particular promise in Legal Question Answering (LQA) \citep{LAI2024181}.
However, unlike general knowledge-based QA \citep{yang-etal-2018-hotpotqa,MultiRC2018,talmor-etal-2019-commonsenseqa},
LQA demands not only substantial domain expertise but also answers that are rigorously grounded in explicit legal authority.
In practice, many legal questions cannot be resolved by a single statutory provision; instead, they require multi-hop reasoning that sequentially integrates multiple legal issues and statutory requirements \citep{zhou2023boosting}.
Such complexity substantially increases the risk of unfaithful or hallucinated reasoning, making it crucial to ensure that generated answers are firmly supported by verifiable legal evidence \citep{Magesh_Hallucination-free}.
Accordingly, accurate retrieval of relevant legal evidence has emerged as a fundamental prerequisite for building reliable LQA systems.

In statutory LQA, the primary supporting evidence is derived from statutory law.
However, statutory provisions are typically written in abstract and normative language and are often organized within complex hierarchical structures.
By contrast, user queries are usually expressed in natural, fact-oriented language, leading to a pronounced vocabulary gap between questions and statutes \citep{Furnas-1987-vocab}.
As a result, directly applying conventional retrieval methods such as BM25 or general-purpose dense retrievers often fails to reliably identify the appropriate legal provisions,
particularly for complex questions that involve multiple interdependent legal issues \citep{zheng2025reasoning}.

To address the reasoning complexity of multi-hop questions in general QA settings, prior multi-hop QA approaches have proposed generating step-wise reasoning processes using the implicit knowledge of LLMs \citep{NEURIPS2022_9d560961}, or decomposing questions into a sequence of sub-questions that are answered iteratively
\citep{press-etal-2023-measuring, trivedi-etal-2023-interleaving}. These methods represent meaningful progress by breaking complex questions into explicit intermediate reasoning steps, instead of attempting to solve them through a single-step inference.
However, in the context of LQA, when decomposed sub-questions remain in natural language form, the vocabulary gap between user questions and statutory text persists, making it difficult to reliably retrieve the statutory provisions corresponding to each sub-question.

More importantly, we observe that the vocabulary gap in statutory LQA is multifaceted, arising not only from surface-level lexical mismatch but also from differences in normative phrasing and statute-level organization.
Relying on a single query reformulation is therefore insufficient to robustly retrieve relevant statutory provisions. Motivated by this insight, we adopt a multi-channel parametric query refinement strategy, in which each sub-question is reformulated into multiple complementary representations that target different retrieval failure modes.

To this end, we propose \textbf{Decompose-and-Refine (\methodName{})}, a framework that tightly integrates question decomposition with parametric knowledge-–based query refinement. \methodName{} progressively decomposes a complex legal question into atomic sub-questions and generates multiple complementary parametric queries for each sub-question based on legal knowledge (Figure \ref{fig:overview}).
This design enables issue-level alignment between decomposed legal questions and their corresponding statutory evidence, allowing the selection of a single core statutory provision for each sub-question. By incrementally accumulating issue-aligned evidence, \methodName{} reduces unnecessary context expansion while supporting evidence-centric legal reasoning.

Experimental results on KoBLEX \citep{lee-etal-2025-koblex},
a Korean multi-hop LQA benchmark grounded in statutory law,
demonstrate that \methodName{} consistently improves both retrieval accuracy and final answer quality. By explicitly separating and accumulating statutes on a per-issue basis, \methodName{} maintains interpretability of the reasoning process while achieving substantial performance gains.

Our contributions are summarized as follows:
\begin{itemize}
    \item We propose \methodName{}, a novel statute-grounded LQA framework that integrates step-wise question decomposition with multi-channel parametric query refinement.
    
    \item We align each atomic sub-question with exactly one supporting statutory provision, enabling interpretable, issue-level legal reasoning and transparent verification. 
    
    \item We demonstrate on the KoBLEX benchmark that \methodName{} consistently outperforms existing baselines in both retrieval and answer accuracy, while relying solely on a standard BM25 retriever without domain-specific fine-tuning.
\end{itemize}
\section{Related Works}
\subsection{Query Refinement}

A vocabulary gap often exists between user queries and target documents \citep{Furnas-1987-vocab}. To mitigate this gap, prior work has explored various query refinement and expansion techniques, ranging from lexical expansion to semantic reformulation \citep{Carpineto-2012-QueryExpansion}.

With the advent of LLM, query refinement has shifted from surface-level transformations toward semantic reformulation.
For example, \citet{gao-etal-2023-precise} generate hypothetical documents corresponding to a given query,
while \citet{wang-etal-2023-query2doc} leverage LLMs to produce pseudo-documents for retrieval expansion.

In the legal domain, query rewriting research has predominantly focused on case law retrieval.
\citet{Askari2021CombiningLA} propose summarizing lengthy case law queries or extracting key terms,
while \citet{zhou2023boosting} improve retrieval performance by using LLMs to select or summarize salient content that is decisive for judicial outcomes.
More recently, \citet{kim-etal-2025-gure} propose a generative query rewriting approach that leverages LLMs to alleviate vocabulary mismatch between queries and legal documents.

However, these approaches primarily target case law retrieval, where the objective is to identify precedents with similar factual circumstances, and thus differ substantially from statute-based LQA, which requires precise alignment with normative legal provisions. Moreover, the applicability of multi-channel query refinement -- in which a single query is transformed into multiple complementary variants for retrieval---has received limited attention in the context of LQA.

\subsection{Legal Question Answering}
Several existing LQA benchmarks \citep{guha2023legalbench, fei-etal-2024-lawbench, fan2025lexam} focus on settings in which LLMs directly generate answers without an explicit retrieval step. While such benchmarks are useful for evaluating the legal knowledge encoded in LLMs,
they do not adequately reflect the evidence retrieval and verification processes required in real-world LQA.
\citet{2024LouisInterpretable} introduce an LQA benchmark and baselines based on a retrieve-and-read paradigm and evaluate multiple LLMs; however, they do not explicitly address issue-level retrieval strategies, limiting their applicability to complex legal questions involving multiple intertwined issues.

Prior multi-hop QA approaches \citep{press-etal-2023-measuring,trivedi-etal-2023-interleaving,jiang-etal-2023-active,cao-etal-2023-probabilistic,chu-etal-2024-beamaggr}
improve complex reasoning through step-wise decomposition and iterative retrieval. However, they generally use decomposed sub-questions directly as retrieval queries, paying limited attention to retrieval-oriented query reformulation. More recently, DualRAG \citep{cheng-etal-2025-dualrag} introduces a dual-process retrieval framework that iteratively combines retrieval and reasoning to improve multi-hop question answering performance, while L-MARS \citep{wang2025mars} studies multi-step legal reasoning with retrieval-augmented generation in the legal domain. While these approaches improve reasoning quality through tighter retrieval-reasoning integration, they primarily focus on iterative evidence acquisition and reasoning generation, rather than explicitly transforming decomposed legal issues into retrieval-effective statute-oriented queries. As a result, evidence corresponding to multiple legal issues may still be retrieved and utilized in a mixed manner.

In the legal domain, vocabulary mismatch between user queries and statutory text occurs frequently, and a single legal issue often spans multiple requirements and conditions distributed across statutory structures \citep{zhou2023boosting}.
As a result, question decomposition alone is insufficient to reliably identify the correct statutory provision corresponding to each sub-question. This highlights that, in LQA, not only how to decompose a question, but also how to transform decomposed sub-questions into retrieval-effective queries, is an equally important challenge. 

Motivated by this observation, \citet{lee-etal-2025-koblex} propose ParSeR that leverages parametric provisions to bridge the representational gap between questions and statutes.
However, ParSeR maintains a single-query formulation for the entire question, which limits its ability to separately handle multiple legal issues within a single query.
Consequently, statutory provisions corresponding to different legal issues may be retrieved and used in a mixed manner,
potentially reducing the transparency of evidence usage in downstream reasoning.

Recent studies further report that, in legal reasoning tasks,
providing more contextual information does not necessarily lead to better performance \citep{kim2025legalaifail}.
Instead, selectively identifying and utilizing only the core statutory provisions relevant to each issue is crucial for improving both accuracy and interpretability.

The proposed \methodName{} integrates question decomposition with parametric query refinement to construct issue-level statute-aligned evidence. By selectively accumulating only core statutory provisions for each legal issue, \methodName{} enables precise and interpretable legal reasoning.

\begin{figure*}[ht]
\begin{center}
   \includegraphics[width=\linewidth]{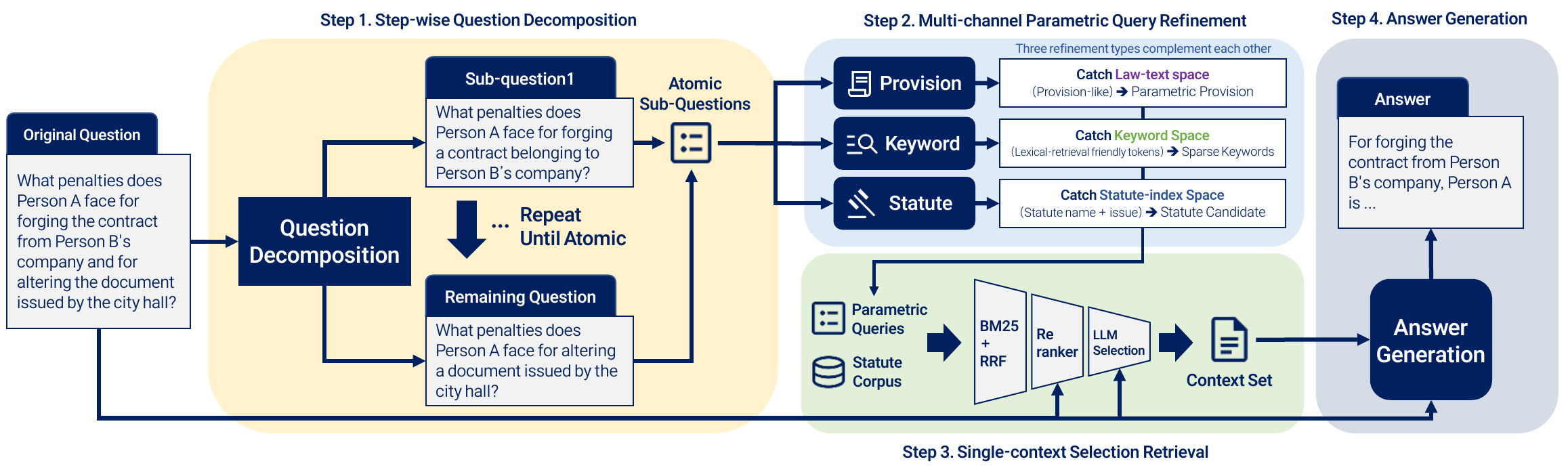}
\end{center}
   \caption{Illustration of Decompose-and-Refine (\methodName) framework. Given a legal question, the model iteratively decomposes it into atomic sub-questions. For each sub-question, \methodName{} generates multiple complementary parametric queries, including provision-style queries, sparse keywords, and candidate statute queries. Each query is used for BM25-based lexical retrieval, and the resulting candidate sets are fused via RRF and reranked, and a single most relevant statute is selected. Selected statutes are accumulated across reasoning steps and used to generate a final context-grounded answer.}
\label{fig:method}
\vspace{-7pt}
\end{figure*}

\section{Method: Decompose-and-Refine}
\label{sec:method}

\methodName{} aims to simultaneously improve retrieval effectiveness and reasoning accuracy by progressively decomposing a complex legal question into atomic sub-questions and generating multiple law-informed query representations for each sub-question.

The overall pipeline consists of four stages:
(1) step-wise question decomposition,
(2) multi-channel parametric query refinement,
(3) single-context selection retrieval, and
(4) answer generation
(Figure~\ref{fig:method}).
Detailed prompt templates used at each stage are provided in Appendix~\ref{appendix:prompt}.

\subsection{Step-wise Question Decomposition}
\label{subsec:decomposition}

Given an input question $Q$, \methodName{} processes it sequentially using a step-wise decomposition strategy.
At each step, the model determines whether the current question is an \textit{atomic question} that cannot be further decomposed, or a \textit{non-atomic question} that requires additional decomposition.
If the question is classified as non-atomic, the model extracts a single atomic sub-question $q_i$ to be resolved at the current step and reconstructs the remaining unresolved part as the next question $q_{i+1}$.
By iteratively applying this procedure, the original question is decomposed into a sequence of sub-questions of up to $T$ steps. The decomposition process terminates when the remaining question $q_{i+1}$ is classified as atomic and treated as the final reasoning target.

\subsection{Multi-channel Parametric Query Refinement}
\label{subsec:parametric}

For each atomic sub-question $q_i$,
\methodName{} generates a \textbf{set of complementary parametric queries} rather than relying on a single query representation. Each refinement type targets a distinct retrieval failure mode and plays a complementary role in bridging the gap between natural-language questions and statutory text.

Specifically, we employ the following three types of query refinement strategies. 

\paragraph{(1) Parametric Provision.}
Given a sub-question $q_i$, the LLM generates a set of sentences
$\{p_{i,1}, \dots, p_{i,k}\}$ formulated to mirror statutory language and structure.
By projecting the query into a law-centric representation space,
this channel addresses linguistic mismatch by aligning the query with the normative phrasing of statutes.

\paragraph{(2) Parametric Sparse Keyword.}
For the same sub-question, the LLM extracts core legal concepts and procedural elements to produce a concise set of 5--10 sparse, noun-centric keywords.
By emphasizing legally salient terms while discarding syntactic structure,
this channel is well suited for lexical-based retrieval methods, which are particularly effective in the legal domain due to the term-sensitive nature of statutory text.

\paragraph{(3) Parametric Statute Candidate.}
Finally, the LLM predicts a set of statute names that are likely to be relevant to the sub-question and generates statute-aware retrieval queries by combining each statute name with the corresponding legal issue.
By restricting retrieval to these predicted statutes, this channel reduces corpus-level ambiguity and suppresses spurious matches.

To make the distinctions among the three refinement strategies more concrete,
we provide a brief illustrative example below.

\textbf{Illustrative Example.}
Given the sub-question:
\textit{``When managing Person A’s property, whose interests should Person B prioritize?''}

\begin{itemize}
    \item \textbf{Parametric Provision:}
    ``A trust is established for the purpose of managing trust property for the benefit of the beneficiary and using and deriving profits from it in accordance with its intended purpose.'' (...)
    
    \item \textbf{Sparse Keyword:}
    trust, beneficiary priority, trustee duty, breach of duty, damages, trust property management, (...)
    
    \item \textbf{Statute Candidate:}
    (1) Civil Act, beneficiary’s interests in a trust (2) Trust Act, trustee’s fiduciary duties (...)
\end{itemize}

\subsection{Single-context Selection Retrieval}
\label{subsec:retrieval}

For each sub-question, we design the retrieval process to yield exactly one supporting statutory provision. First, we perform BM25-based lexical retrieval \citep{Robertson2009BM25} for each parametric query generated in Section \ref{subsec:parametric}. We observe that statutory provisions appearing repeatedly across retrieval results
from multiple parametric queries are more likely to be semantically relevant to the underlying sub-question.
Based on this intuition, we aggregate the retrieval results using Reciprocal Rank Fusion (RRF) \citep{rrf_cormack2009},
which assigns higher ranks to provisions consistently retrieved by different queries. Following \citet{lee-etal-2025-koblex}, we then apply a cross-encoder–based reranker
to the top $N$ candidate provisions obtained from RRF,
re-ranking them according to their semantic relevance to the sub-question.

Finally, given the re-ranked candidate set, an LLM selection module chooses a single statutory provision that most directly answers the corresponding sub-question. Through this process, \methodName{} retains only the minimal core evidence required for each legal issue. We further analyze the effect of enforcing single-statute selection for each decomposed sub-issue in Appendix~\ref{appendix:statute_selection_ablation}.

\subsection{Answer Generation}
\label{subsec:qa}

The statutory provisions selected at each step are accumulated iteratively, forming a deduplicated evidence set.
In the final stage, the original question $Q$ and the accumulated statutory provisions are provided as input to a context-grounded QA model, which generates the final answer based solely on the given legal evidence.
We adopt a standard prompting strategy, in which the model is given the retrieved statutory context together with a small number of few-shot examples and is directly prompted to generate the answer.

Overall, by integrating step-wise question decomposition with parametric query refinement, \methodName{} suppresses unnecessary retrieval expansion while systematically constructing an interpretable, issue-aligned set of statutory evidence.
Given that \methodName{} already yields tightly grounded, issue-specific contexts, we hypothesize that direct answer generation is preferable to explicit reasoning narration, a claim we empirically validate in Appendix~\ref{appendix:answer_prompting_ablation}.

\section{Experiments}
\label{sec:experiments}

This section describes the experimental setup used to evaluate the effectiveness of \methodName{}.

All experiments are conducted in a few-shot setting,
where pre-trained large language models are used directly without any additional task-specific training.
We employ Qwen3-32B \citep{yang2025qwen3technicalreport} and Gemma3-27B \citep{team2025gemma}, both of which support multilingual QA, as the backbone LLMs.
For retrieval, we use a standard BM25 retriever \citep{Robertson2009BM25}, and a BGE-based reranker \citep{chen-etal-2024-m3} is applied for candidate re-ranking.
Further experimental details are provided in Appendix~\ref{appendix:experimental_details}.

\begin{table}[t]
\centering
\resizebox{\columnwidth}{!}{%
\begin{tabular}{lcc}
\toprule
Method & Decomp. & QR. \\
\hline
Standard / CoT & $\times$ & $\times$ \\

\makecell[l]{Self-Ask / IRCoT / \\ FLARE / ProbTree / BeamAggr}
& $\checkmark$ & $\times$ \\

ParSeR & $\times$ & $\checkmark$ \\ \midrule
\textbf{\methodName{}} (Ours) & $\checkmark$ & $\checkmark$ \\
\bottomrule
\end{tabular}%
}
\caption{Comparison of baseline methods along two key design dimensions. Decomp. and QR indicate
\textit{step-wise question decomposition} and
\textit{query refinement}.}
\label{tab:design_axes}
\vspace{-9pt}
\end{table}

\begin{table*}[ht]
\resizebox{\linewidth}{!}{%
\begin{tabular}{ccccccccc}
\toprule
 \multicolumn{1}{l}{} & \multicolumn{4}{c}{Retrieval} & \multicolumn{4}{c}{Answer} \\ \hline
 & \multicolumn{2}{c}{F-1} & \multicolumn{2}{c}{EM} & \multicolumn{2}{c}{Token F-1} & \multicolumn{2}{c}{LF-Eval} \\ \hline
 & Qwen & Gemma & Qwen & Gemma & Qwen & Gemma & Qwen & Gemma \\ \hline
SP \citep{NEURIPS2020_1457c0d6} & - & - & - & - & 31.77 & 31.29 & 42.89 & 45.54 \\
CoT \citep{NEURIPS2022_9d560961} & - & - & - & - & 26.78 & 28.51 & 41.99 & 48.01 \\
Self-Ask \citep{press-etal-2023-measuring} & 31.62 & 33.64 & 13.27 & 8.41 & 33.03 & 19.22 & 50.52 & 36.96 \\
IRCoT \citep{trivedi-etal-2023-interleaving} & 29.65 & 17.55 & 4.42 & 2.65 & 29.86 & 22.98 & 49.36 & 33.95 \\
FLARE \citep{jiang-etal-2023-active} & 14.94 & 15.93 & 3.1 & 2.65 & 30.7 & 27.45 & 46.04 & 44.27 \\
ProbTree \citep{cao-etal-2023-probabilistic} & 23.03 & 21.74 & 6.64 & 4.87 & 27.25 & 28.06 & 50.67 & 49.96 \\
BeamAggr \citep{chu-etal-2024-beamaggr} & 21.08 & 21.79 & 5.75 & 1.77 & 31.32 &27.64 & 44.05 & 45.70 \\
ParSeR \citep{lee-etal-2025-koblex} & 46.24 & 45.27 & 17.70 & 13.27 & 40.65 & 42.29 & 56.00 & 61.16 \\ \midrule
\textbf{\methodName{} (ours)} & \textbf{58.63} & \textbf{55.08} & \textbf{34.51} & \textbf{28.76} & \textbf{45.53} & \textbf{43.97} & \textbf{68.13} & \textbf{65.35} \\ \bottomrule
\end{tabular}%
}
\caption{Experimental Results on KoBLEX benchmark. The table compares retrieval performance (F-1 / EM) and answer quality (Token F-1 / LF-Eval) under Qwen3-32B and Gemma-27B. DAR (ours) consistently outperforms prior reasoning and retrieval-based baselines, demonstrating the effectiveness of query refinement and structured retrieval.}
\label{tab:main_result}
\end{table*}

\subsection{Dataset and Evaluation Metrics}
\label{subsec:dataset}

We conduct experiments on KoBLEX \citep{lee-etal-2025-koblex},
a statutory law–grounded multi-hop legal question answering benchmark. KoBLEX consists of 226 question–answer pairs,
each of which requires sequential reasoning over multiple legal issues and their corresponding statutory provisions.
Each sample includes a \textit{background scenario}, a \textit{question}, an \textit{answer}, a set of gold statutory provision contexts required to derive the answer, and a corpus of statutory provisions serving as the retrieval collection.

We evaluate both answer quality and retrieval effectiveness.
\begin{itemize}
    \item \textbf{Answer Quality.}
    We measure answer quality using token-level F-1 score
    and LF-Eval, an LLM-as-a-judge metric designed to assess legal faithfulness.
    \item \textbf{Retrieval Quality.}
    To evaluate retrieval performance, we compute provision-level Exact Match (EM) and F-1 scores between the set of provisions selected by the model and the gold provision set.
\end{itemize}

\subsection{Baselines}
\label{subsec:baselines}

To assess the performance of \methodName{}, we compare it against a diverse set of baseline methods, which are categorized along two design axes: whether question decomposition is employed and whether parametric query refinement is used (Table~\ref{tab:design_axes}).

\begin{itemize}
    \item \textbf{Direct QA Methods.}
    Methods that process the entire question as a single input
    and generate answers end-to-end
    without explicit question decomposition
    or parametric query refinement
    (Standard Prompting \citep{NEURIPS2020_1457c0d6},
    Chain-of-Thought \citep{NEURIPS2022_9d560961} (CoT)).

    \item \textbf{Question Decomposition Methods.}
    Methods that decompose a question into sub-questions or intermediate reasoning steps
    to enable step-wise reasoning and retrieval,
    but do not explicitly incorporate parametric query refinement
    (Self-Ask \citep{press-etal-2023-measuring},
    IRCoT \citep{trivedi-etal-2023-interleaving},
    FLARE \citep{jiang-etal-2023-active},
    ProbTree \citep{cao-etal-2023-probabilistic},
    BeamAggr \citep{chu-etal-2024-beamaggr}).

    \item \textbf{Parametric Query Reformulation Methods.}
    Methods that rewrite the question into parametric representations
    aligned with statutory language
    to improve retrieval performance,
    but do not perform explicit question decomposition
    (ParSeR \citep{lee-etal-2025-koblex}).
\end{itemize}

All baseline methods are implemented using the same statutory corpus and an identical retrieval pipeline to ensure a fair comparison. Specifically, for all retrieval-based methods, we apply BM25-based lexical retrieval followed by cross-encoder reranking and an LLM-based selection module, using the same hyperparameter settings as in \methodName{}.
Direct QA methods, including Standard Prompting and CoT, do not involve any retrieval step and generate answers solely based on the input question.
To further control for model capacity, the same backbone LLMs are used across all methods, including \methodName{}.
Each baseline follows the original experimental settings proposed in the corresponding literature, with only minimal adaptations to accommodate the legal domain.

\section{Results and Analysis}

\subsection{Overall Results}

Table~\ref{tab:main_result} presents the overall experimental results on the KoBLEX benchmark.
\methodName{} substantially outperforms all baseline methods in both retrieval effectiveness and final answer quality across both backbone LLMs (Qwen3-32B and Gemma3-27B).

In particular, in terms of retrieval F-1, \methodName{} achieves substantial gains, approaching a twofold improvement relative to existing question decomposition–based methods (Self-Ask, IRCoT, FLARE, ProbTree, and BeamAggr), and also records the highest accuracy under the EM metric. These results indicate that question decomposition alone is insufficient for effective statutory retrieval, and that reconstructing decomposed questions into law-aligned representations is essential for identifying the correct legal evidence.

\methodName{} also consistently outperforms ParSeR,
demonstrating the advantage of explicitly handling multiple legal issues in a step-wise manner while preserving the benefits of parametric query refinement. This suggests that combining issue-level decomposition with parametric reformulation is more effective than applying either strategy in isolation.

From a design perspective (Table~\ref{tab:design_axes}), Question decomposition–based methods provide a step-wise reasoning structure, but fail to sufficiently address the vocabulary gap because decomposed sub-questions are used directly as retrieval queries. ParSeR improves retrieval through parametric provisions, but its single-query formulation leads to the mixing of evidence when multiple legal issues are present within a single question.

By integrating these two design axes, \methodName{} enforces the selection of exactly one core statutory provision per sub-question. This design choice leads to consistent improvements in both retrieval accuracy and reasoning transparency.

\begin{table}[t]
\centering
\resizebox{\columnwidth}{!}{%
\small
\begin{tabular}{ccc}
\toprule
Decomp. & QR. & F-1 / EM \\
\midrule
$\checkmark$ 
& Par. $+$ Spar. $+$ Stat. 
& \textbf{58.63 / 34.51} \\

$\checkmark$
& Par. $+$ Spar. 
& 57.73 / 33.19 \\

$\checkmark$
& Par. 
& 55.52 / 30.09 \\

$\checkmark$
& $\times$
& 45.44 / 26.55 \\

$\times$
& Par. 
& 46.74 / 19.03 \\
\bottomrule
\end{tabular}%
}
\caption{Ablation study of \methodName{} on retrieval performance (Qwen3-32B).
Par., Spar., and Stat. denote \textit{parametric provision} queries,
\textit{parametric sparse keyword} queries, and
\textit{parametric statute candidate} queries, respectively,
while Decomp. and QR indicate
\textit{step-wise question decomposition} and
\textit{query refinement}.}
\label{tab:ablation}
\end{table}

\begin{figure}[t]
    \centering
    \includegraphics[width=\linewidth]{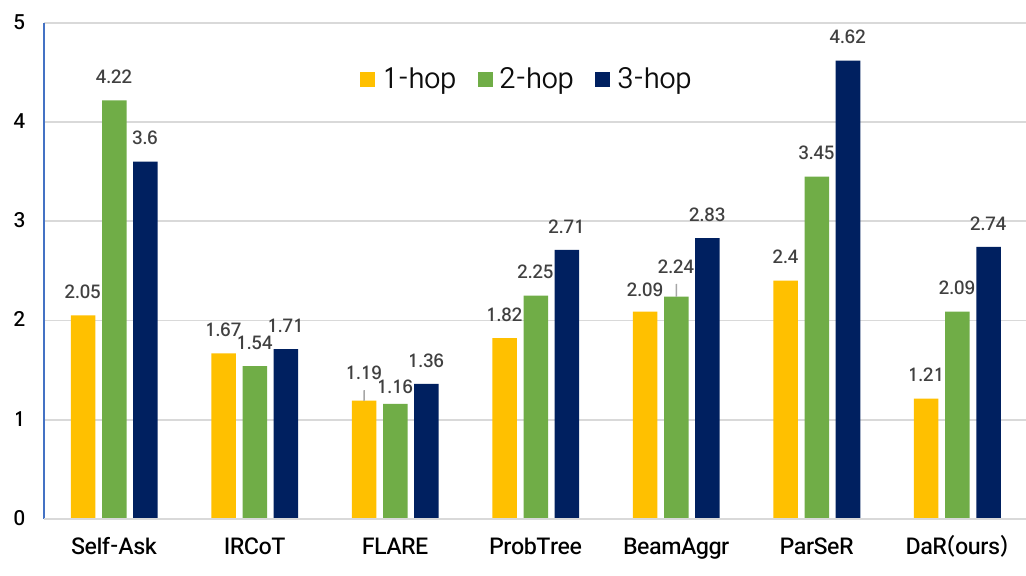}
    \caption{Average number of retrieved statutes per question
    across different number of gold contexts.
    \methodName{} exhibits a controlled and near-linear increase in retrieved evidence,
    closely aligned with the underlying evidence requirements of the questions.}
    \label{fig:avg_retrieved_statutes}
\end{figure}
\subsection{Ablation Study: Components}

Table~\ref{tab:ablation} reports the results of an ablation study examining the contributions of the main components of \methodName{}.

We first analyze the effect of parametric query refinement.
When all three types of parametric queries
(provision, sparse keyword, and statute candidate) are employed, \methodName{} achieves the highest retrieval performance, with F-1/EM scores of 58.63/34.51. In contrast, removing query refinement entirely (i.e., using question decomposition alone) results in a substantial performance drop to 45.44/26.55. This confirms that parametric query refinement
is a core component supporting retrieval effectiveness.

Moreover, a gradual performance degradation is observed
as individual query types are removed: removing statute-aware queries yields 57.73/33.19, while removing both statute-aware and sparse keyword queries further reduces performance to 55.52/30.09. These trends indicate that the different query types contribute complementarily to retrieval performance.

The importance of question decomposition is even more pronounced. Compared to the configuration that retains decomposition while applying parametric query refinement (55.52/30.09), removing decomposition and relying solely on parametric queries leads to a sharp drop in retrieval performance, with F-1 decreasing to 46.74 and EM to 19.03.
This result demonstrates that query refinement alone is insufficient for accurately organizing and aligning multiple legal issues within a complex question.

Overall, these findings suggest that in statutory LQA,
query refinement by itself is not enough; explicit question decomposition is essential for structurally isolating and ordering legal issues. By integrating decomposition with parametric query refinement, \methodName{} reliably identifies the core statutory provision for each issue, leading to simultaneous improvements in both retrieval effectiveness and final answer quality.

\subsection{Controlled Evidence Accumulation}

Figure~\ref{fig:avg_retrieved_statutes} reports the average number of retrieved statutes per question, grouped by the number of gold contexts and method.
This analysis highlights how different methods scale their use of evidence as the number of required supporting statutes increases.

Across existing baselines, the number of retrieved statutes does not exhibit a consistent relationship with the number of gold contexts.
For question decomposition–based methods such as Self-Ask, IRCoT, FLARE, ProbTree, and BeamAggr, the amount of retrieved evidence fluctuates across different levels of gold contexts and does not increase monotonically as more supporting statutes are required. ParSeR retrieves a relatively large number of statutes across all settings, but the growth is not well aligned with the underlying multi-context structure of the questions.

In contrast, \methodName{} shows a clear and gradual increase in the number of retrieved statutes as the required gold contexts increases (1.21 for questions with 1-hop, 2.08 for 2-hop, and 2.74 for 3-hop).
This pattern closely reflects the actual evidence requirements of the questions, indicating that \methodName{} retrieves statutes in a structured manner. By aligning evidence accumulation with the number of required gold contexts,
\methodName{} avoids both under-retrieval and unnecessary context expansion, supporting precise and interpretable multi-hop legal reasoning.

\begin{table}[t]
\centering
\small
\resizebox{\columnwidth}{!}{%
\begin{tabular}{lcc}
\toprule
Retriever Setting & F-1 & EM \\ \midrule
Sparse (with Query Refinement) & 58.63 & 34.51 \\
Dense (with Query Refinement) & 58.20 & 34.07 \\
Sparse (without Query Refinement) & 45.44 & 26.55 \\
Dense (without Query Refinement) & 46.46 & 27.43 \\ \bottomrule
\end{tabular}%
}
\caption{Retrieval performance comparison between sparse (BM25) and dense (BGE) retrievers under a fixed question decomposition setting, with and without query refinement.}
\label{tab:ablation_retriever}
\end{table}

\subsection{Ablation Study: Retriever Type}

Table~\ref{tab:ablation_retriever} compares the retrieval performance of sparse and dense retrievers under a fixed question decomposition setting, with and without query refinement.
A common intuition is that dense retrievers, by capturing semantic similarity in embedding space, may reduce the need for explicit query refinement.
To examine this hypothesis, we performed additional experiments using a Korean fine-tuned dense retriever based on the BGE embedding model \citep{chen-etal-2024-m3}\footnote{\url{https://huggingface.co/dragonkue/BGE-m3-ko}} and compared it with a standard BM25-based sparse retriever, while keeping the other procedure identical in all settings.

The results show that when query refinement is applied, the performance difference between sparse and dense retrievers becomes relatively small.
Sparse retrieval with query refinement achieves an F-1/EM of 58.63/34.51, while dense retrieval with the same refinement yields 58.20/34.07.
This suggests that, under structured question decomposition and law-aligned query refinement, a sparse lexical retriever can achieve retrieval performance comparable to that of the evaluated dense retriever.

In contrast, when query refinement is removed, retrieval performance drops substantially for both retriever types.
Dense retrieval without query refinement achieves an F-1/EM of 46.46/27.43, showing a slight improvement over sparse retrieval without refinement (45.44/26.55).
However, the improvement remains limited, suggesting that retriever architecture alone may not adequately resolve complex legal questions involving multiple legal issues and fine-grained statutory conditions.

Overall, these findings suggest that the performance gains of \methodName{} are driven more by explicit query refinement and issue-level query construction than by retriever architecture alone.
Notably, \methodName{} achieves strong retrieval performance using a standard BM25 retriever without additional dense retriever fine-tuning, highlighting the practicality of the proposed framework.
\section{Conclusion}

In this paper, we presented \textbf{Decompose-and-Refine (DAR)}, a statute-grounded framework for legal question answering that targets the challenges of multi-hop statutory reasoning.
By integrating step-wise question decomposition with parametric query refinement, \methodName{} enables effective alignment between natural-language questions and normative statutory text and demonstrates strong performance on the KoBLEX benchmark, even without retriever fine-tuning.
We further demonstrate that \methodName{} supports controlled and interpretable evidence accumulation by selecting a minimal, issue-aligned set of statutory provisions, thereby facilitating transparent and reliable legal reasoning.

\section*{Limitations}

Despite its effectiveness on statute-grounded multi-hop legal question answering, the proposed \methodName{} has several limitations.

First, \methodName{} is specifically designed for statutory law–grounded legal question answering, and its applicability to case law–centric reasoning remains unexplored. Statute-based LQA requires precise alignment with normative provisions, whereas case law retrieval focuses on identifying precedents with similar factual patterns and judicial reasoning. As a result, the query structures, evidence granularity, and relevance criteria differ substantially between the two settings. Extending \methodName{} to case law would therefore require alternative query reformulation strategies and retrieval objectives tailored to precedent-based reasoning.

Second, \methodName{} relies heavily on the accuracy of step-wise question decomposition. When the decomposition stage fails to correctly isolate the core legal issues, errors can propagate to subsequent parametric query generation, retrieval, and answer generation stages. As illustrated in Appendix~\ref{appendix:case_study}, an incorrect or misaligned sub-question can lead to systematically irrelevant statute retrieval, even when later components function as intended. This error propagation highlights the importance of robust decomposition mechanisms and suggests that incorporating decomposition verification or correction strategies may be a promising direction for future work.

Overall, these limitations point to future research directions involving broader legal domains, more resilient decomposition techniques, and adaptive evidence modeling beyond strictly statute-grounded settings.

\section*{Ethical Considerations}

The KoBLEX dataset is released under the Creative Commons Attribution--NonCommercial 4.0 International (CC BY-NC 4.0) license.\footnote{\url{https://github.com/daehuikim/KoBLEX}} 
All question--answer pairs and supporting statutory contexts were constructed based solely on publicly available statutory law and were further reviewed and manually revised by legal experts. As a result, the dataset does not contain personal data, private information, or records of real individuals, and thus does not pose risks related to privacy infringement or data misuse.

All experiments in this study were conducted using LLMs with permissive or explicitly stated licenses, including Qwen3, which is released under the Apache 2.0 license,\footnote{\url{https://github.com/QwenLM/Qwen3}} and Gemma 3 with its own license.\footnote{\url{https://ai.google.dev/gemma/terms}} 

In preparing this manuscript, we also made limited use of AI-assisted writing tools (GPT-5) to translate Korean examples into English. These translations were reviewed by the authors to ensure faithfulness.

The intended use of this research is to support academic investigation into LQA methods and explainable multi-hop reasoning over statutory law. The proposed framework are not designed for deployment in real-world legal practice, nor should they be interpreted as providing legal advice or guidance for resolving actual disputes or incidents.

%\section*{Acknowledgments}

\bibliography{custom}

\appendix

\section{Experimental Details}
\label{appendix:experimental_details}

This section describes the evaluation metrics and the implementation details of the baselines used in our experiments.

\subsection{Implementation Details}
\label{appendix:retreival_details}

In the overall pipeline of \methodName{}, all major components—including question decomposition, parametric provision generation, provision selection, and final answer generation —are implemented using the same LLM.
This design choice eliminates model-induced variance across different stages and allows us to more clearly analyze the effectiveness of the proposed method itself.

For each atomic sub-question, all generated parametric queries are used as inputs to a BM25-based lexical retriever \citep{Robertson2009BM25}, retrieving the top 100 statutory provisions per query as candidates. Retrieval results obtained from different parametric queries are then aggregated using Reciprocal Rank Fusion (RRF) \citep{rrf_cormack2009}, with the RRF constant set to 60. Based on the aggregated RRF scores, the top 300 candidate provisions are selected and passed to a cross-encoder–based reranker. After reranking, only the top 10 provisions are forwarded to the selection module.
For reranking, we use a Korean fine-tuned BGE reranker \citep{chen-etal-2024-m3}\footnote{\url{https://huggingface.co/dragonkue/bge-reranker-v2-m3-ko}}.

All experiments are conducted on a single NVIDIA A100 GPU.
Inference over the full KoBLEX benchmark takes approximately three hours.

\subsection{Evaluation Metrics}
\label{subsec:metric_details}

To evaluate the performance of \methodName{} from both answer accuracy and legal grounding perspectives, we employ multiple complementary evaluation metrics.

\paragraph{Retrieval F-1.}
Retrieval F-1 measures the degree of partial overlap between
the set of statutory provisions selected by the model
and the gold-standard provision set.
For each question, precision and recall are computed based on the selected and gold provisions, and their harmonic mean is reported as the F-1 score.

\paragraph{Retrieval Exact Match (EM).}
Retrieval EM evaluates whether the provision set selected by the model exactly matches the gold-standard provision set.
For each question, the score is set to 1 if the two sets are identical, and 0 otherwise.

\paragraph{Answer Token-level F-1.}
Answer Token-level F-1 measures the token-level overlap
between the generated answer and the gold answer. Both answers are tokenized, and precision and recall are computed based on overlapping tokens, from which the F-1 score is derived.
This metric captures both content accuracy and coverage of key information.

\paragraph{LF-Eval.}
LF-Eval is a legal fidelity–oriented evaluation metric proposed by \citet{lee-etal-2025-koblex}, built upon the G-Eval framework \citep{liu-etal-2023-g}.
Beyond surface-level answer matching, LF-Eval assesses the legal validity and faithfulness of an answer with respect to the provided statutory provisions.
Specifically, an evaluation LLM receives both the generated answer and the cited provisions as input and judges whether the answer can be legally justified in context.
This allows the metric to distinguish between answers that are lexically similar but legally incorrect, and those that are phrased differently yet legally sound. LF-Eval achieves a strong correlation with human judgments, reporting a Pearson correlation of 84.90 in a human evaluation study, which demonstrates that LF-Eval reliably reflects expert assessments of legal correctness.

\subsection{Baseline Implementations}
\label{subsec:baseline_details}

This section describes the implementation details of the baseline methods used in our experiments. Each baseline follows the original configuration proposed in prior work as closely as possible, with only minimal modifications to ensure applicability to the legal domain and the KoBLEX dataset.

\paragraph{Direct QA Methods.}
\begin{itemize}
    \item \textbf{Standard Prompting} \citep{NEURIPS2020_1457c0d6}
    generates answers directly from the LLM without using any retrieval module, by providing the question together with a small number of few-shot examples.

    \item \textbf{Chain-of-Thought (CoT)} \citep{NEURIPS2022_9d560961}
    uses few-shot prompts that explicitly demonstrate step-by-step reasoning, encouraging the model to generate intermediate reasoning chains before producing the final answer, without external retrieval.
\end{itemize}

\paragraph{Question Decomposition Methods.}
\begin{itemize}
    \item \textbf{Self-Ask} \citep{press-etal-2023-measuring}
    decomposes a complex question into a sequence of follow-up questions and generates intermediate answers sequentially.
    In our implementation, each follow-up question is processed using the same three-step retrieval pipeline (BM25, reranking, and selection) to select a single supporting statute before generating the intermediate answer.

    \item \textbf{IRCoT} \citep{trivedi-etal-2023-interleaving}
    alternates between CoT reasoning and retrieval. 
    Each retrieval step is implemented using the unified three-step retrieval pipeline, and intermediate reasoning outputs are used as queries for subsequent iterations.

    \item \textbf{FLARE} \citep{jiang-etal-2023-active}
    monitors token-level generation confidence and triggers retrieval when low-confidence segments are detected.
    Whenever retrieval is triggered, the same three-step retrieval pipeline is applied using the partial generation as a query.

    \item \textbf{ProbTree} \citep{cao-etal-2023-probabilistic}
    decomposes a question into a hierarchical tree and evaluates multiple reasoning strategies at each node.
    For open-book strategies, retrieval is performed using the unified three-step retrieval pipeline before computing log-probability scores.

    \item \textbf{BeamAggr} \citep{chu-etal-2024-beamaggr}
    aggregates multiple candidate answers generated at lower-level nodes using beam search.
    Retrieval at each node is implemented via the same three-step retrieval pipeline, ensuring consistency across methods.
\end{itemize}

\paragraph{Parametric Retrieval Methods.}
\begin{itemize}
    \item \textbf{ParSeR} \citep{lee-etal-2025-koblex}
    reformulates a question into statute-like parametric queries to improve retrieval effectiveness.
    Since an official implementation of ParSeR tailored to the KoBLEX benchmark is publicly available, we directly use the authors’ official code\footnote{\url{https://github.com/daehuikim/KoBLEX}} without modification. The same three-step retrieval pipeline is applied to ParSeR.
\end{itemize}
\section{Prompt Design}
\label{appendix:prompt}

This section presents the prompts used at each stage of \methodName{}, as described in Section~\ref{sec:method}.
Several prompts are adopted directly from prior work without modification. Specifically, the Parametric Provision Generation Prompt, the Selection Prompt, and the Answer Generation Prompt are reused from ParSeR~\citep{lee-etal-2025-koblex}.

For each prompt, we describe its intended purpose and usage context. Slots where actual inputs are injected are indicated in \textit{italic}, and illustrative examples are omitted for brevity.

\begin{tcolorbox}[title=Question Decomposition Prompt, colback=white, colframe=black!40, breakable]
\footnotesize
\textbf{System:} You are a legal reasoning assistant for Korean law.
Your task is to decompose a complex legal question into smaller, atomic sub-questions one step at a time, and update the remaining question.\
\\
Definitions:\\
- Atomic sub-question: a question that can be answered by referring to a small number of specific legal provisions, without further decomposition.\\
- Non-atomic question: a question that still consists of multiple legal issues, steps, or conditions.
\\\\
At each step:\\
1. Read the Background and the CURRENT Question.\\
2. Decide whether the CURRENT Question is atomic or not.\\
3. If NON-ATOMIC:\\
   - Extract EXACTLY ONE atomic sub-question that should be solved at this step.\\
   - Rewrite the remaining unresolved part as the Remaining Question.\\
4. If ATOMIC:\\
   - Treat the CURRENT Question itself as the final Sub-question.\\
   - Set Remaining Question to NONE.\\

Guidelines:\\
- Work in Korean when writing questions.\\
- Keep the sub-question as narrow and concrete as possible.\\
- The Remaining Question + the solved Sub-question should still be logically equivalent to the original question.\\
- If the CURRENT Question is already focused on a single legal issue, mark it as ATOMIC and reuse it as Sub-question.\\

Output format (must strictly follow):\\
\\
Decomposition\\
Type: NON-ATOMIC or ATOMIC\\
Sub-question: <ONE atomic sub-question in Korean>\\
Remaining Question: <Updated remaining question in Korean, or 'NONE' if nothing remains>\\

\textbf{User:} <EXAMPLE> ... ...\\
<Query> \\
Background: \textit{background}\\
Question: \textit{question}\\
\\
Decomposition
\end{tcolorbox}

\begin{tcolorbox}[title=Parametric Provision Generation Prompt, colback=white, colframe=black!40, breakable]
\footnotesize
\textbf{System:} You are an expert legal assistant whose task is to identify and return all relevant statutory provisions that support the answer to a given legal question.\\
Your role is not to provide interpretations, summaries, or conclusions, but to retrieve and list the exact legal clauses that serve as a legal basis for the scenario described.\\
- Answer must be a list of clauses in the following format: ["Name of Law (Title or Clause Summary) Exact clause text or its key portion.","Name of Law (Title or Clause Summary) Exact clause text or its key portion.",...] without any other explanations.\\ 
- Do not provide legal analysis or paraphrased answers—your response must consist solely of statutory provisions.\\
- Do not include article or clause numbers (e.g., "Article 123", "제116조의2")—summarize the clause title briefly in parentheses instead.\\
- If multiple laws are involved, list all clauses together in a single list.\\
- All internal quotation marks in the output must be escaped (i.e., use \", \').\\
- If no directly applicable statutory provision exists, generate the most plausible clause in the same format, as if it were part of the relevant law.\\
- Do not explain or consider actual application, just answer in format.

\textbf{User:} <EXAMPLE> ... ...\\
<Query>\\
Question: \textit{background} $+$ \textit{question}\\
Answer:
\end{tcolorbox}

\begin{tcolorbox}[title=Parametric Sparse Keyword Generation Prompt, colback=white, colframe=black!40, breakable]
\footnotesize
\textbf{System:} You are an expert legal assistant whose task is to rewrite queries for Korean statutory retrieval.\\
Your goal is to convert a user's legal question into a high-performance BM25 sparse keyword query.\\
Maximize the recall of relevant statutory provisions by extracting core legal concepts and converting them into precise keywords.\\

Rules:\\
- Output MUST be a single line of 5 to 10 Korean keywords (space-separated).\\
- Prefer nouns, legal terms, and statute-style terminology.\\
- Strictly REMOVE filler words, particles (조사), verb endings (어미), politeness, and full sentences.\\
- Do NOT include quotes, bullets, numbering, or extra explanations.
- Avoid duplicates; keep keywords diverse and informative.\\
- If the question implies a legal effect (e.g., punishment, cancellation, damages), include the specific legal remedy/effect keyword (e.g., 처벌, 무효, 취소, 해제, 손해배상, 기산점).\\
- If parties/roles matter, include role keywords (e.g., 피고인, 피해자, 임차인, 임대인).\\
- If procedure matters, include procedure keywords (e.g., 항소심, 소송, 재결, 기소).\\

\textbf{User:} <EXAMPLE> ... ...\\
<Query>\\
Background: \textit{background}\\
Question:  \textit{question}\\
Answer:
\end{tcolorbox}

\begin{tcolorbox}[title=Parametric Statute Candidate Generation Prompt, colback=white, colframe=black!40, breakable]
\footnotesize
\textbf{System:} You are an expert legal assistant whose task is to predict the most relevant Korean statute names for the given sub-question, and generate short retrieval queries by combining each statute name with key legal issues. \\

Rules:\\
- Answer must be a Python list of strings.\\
- Format: ["Statute Name Keyword1 Keyword2", "Statute Name Keyword3 Keyword4", ...]\\
- Do not provide any explanations, just the list.\\
- Do not include specific article numbers (e.g., NO "Article 123", NO "제116조의2").\\
- If multiple laws are potentially relevant, generate queries for all of them.\\
- If no directly applicable statute is obvious, infer the most plausible area of law (e.g., Civil Act, Criminal Act).\\
- Keywords should be concise (2~5 words), noun-heavy, and strictly relevant to the statute.\\
- All internal quotation marks must be escaped.\\
- Output MUST be in Korean.\\

\textbf{User:} <EXAMPLE> ... ...\\
<Query>\\
Background: \textit{background}\\
Question:  \textit{question}\\
Answer:
\end{tcolorbox}

\begin{tcolorbox}[title=Selection Prompt, colback=white, colframe=black!40, breakable]
\footnotesize
\textbf{System:} You are given a question, and a list of candidate passages with associated passage IDs.  \\
Your task is to identify the most proper passage that directly support the answer to the question.  \\
Please select only one ID among given candidate passages.\\

**Important instructions:**  \\ 
- Do not include other information except the passage ID in your answer.\\
- Even if multiple passages seem relevant or none seem perfectly appropriate, you must select exactly one passage ID.\\
- your answer should be following after "Answer:" without other explanations. \\
\textbf{User:} <EXAMPLE> ... ...\\
<Query>\\
Background: \textit{background}\\
Question:  \textit{question}\\
Candidate: \textit{canidates} \\
Answer:
\end{tcolorbox}

\begin{tcolorbox}[title=Answer Generation Prompt, colback=white, colframe=black!40, breakable]
\footnotesize
\textbf{System:} You are a legal assistant AI. Given a user Background and a Question, generate a concise and accurate answer by grounding your response only in the provided legal passages (Context). \\
-Do not draw on any external knowledge, statutes, precedent, or opinion.\\
-Keep your answers concise and strictly grounded in the quoted text
-Do not refuse or say insufficient context. Instead, use the provided Context to the fullest and deliver a complete, concise answer grounded solely in those passages.
\textbf{User:} <EXAMPLE> ... ...\\
<Query>\\
Question: \textit{background} $+$ \textit{question}\\
Context:  \textit{context}\\
Answer:
\end{tcolorbox}
\section{Issue-Level Statute Selection Ablation}
\label{appendix:statute_selection_ablation}
We assume that each atomic sub-issue obtained through decomposition can be resolved with a single statutory provision. To examine the validity of this assumption, we conduct an additional ablation study in which each sub-issue is allowed to reference multiple statutes simultaneously. Specifically, we relax the one-statute-per-issue constraint and allow the model to select up to $k \leq 3$ and $k \leq 5$ statutes for each issue decomposition step, while keeping all other experimental settings identical (Qwen3-32B backbone). The results are reported in Table \ref{tab:statute_ablation}.

We observe a consistent degradation across all evaluation metrics as k increases. In particular, Retrieval F1 decreases from 58.63 to 52.32, while Retrieval EM drops substantially from 34.51 to 19.47 when allowing up to five statutes. Similar trends are also observed in answer generation metrics (Token-F1 and LF-Eval).

These findings suggest that, once a legal question is decomposed into fine-grained atomic sub-issues, precise alignment between each sub-issue and its corresponding statutory provision becomes more critical than broad evidence aggregation. Expanding the evidence set with multiple statutes introduces substantial contextual noise and weakens issue-level grounding, ultimately reducing retrieval and answer precision. Overall, the results empirically support our design choice of enforcing single-statute selection per sub-issue as an effective form of controlled evidence regularization.

\begin{table}[t]
\small
\centering
\resizebox{\columnwidth}{!}{
\begin{tabular}{lcccc}
\toprule
                & \multicolumn{2}{c}{Retrieval} & \multicolumn{2}{c}{Answer} \\ \midrule
Setting         & F-1           & EM            & Token F-1     & LF-Eval    \\ \midrule
$k = 1$ (Default) & 58.63         & 34.51         & 45.53         & 68.13      \\
$k \leq 3$           & 53.32         & 21.68         & 44.57         & 67.76      \\
$k \leq 5$           & 52.32         & 19.47         & 43.94         & 67.26      \\ \bottomrule
\end{tabular}}
\caption{Ablation study on the maximum number of statutes selected per sub-issue. Increasing the number of selected statutes consistently degrades both retrieval and downstream answer generation performance.}
\label{tab:statute_ablation}
\end{table}
\section{Answer Prompting Ablation}
\label{appendix:answer_prompting_ablation}

\begin{table}[h]
\centering
\small
\begin{tabular}{lcc}
\toprule
Prompting Strategy & Token F-1  & LF-Eval  \\
\midrule
Naive (Standard Prompting) & \textbf{45.86} & \textbf{67.85} \\
Chain-of-Thought (CoT) & 39.76 & 60.88 \\
CoT Self-Consistency (n=5) & 40.42 & 63.73 \\
\bottomrule
\end{tabular}
\caption{Ablation of answer-generation prompting strategies on KoBLEX (Qwen3-32B).
All settings use the same retrieved contexts; only the final QA prompting differs.}
\label{tab:answer_prompting_ablation}
\end{table}

Table~\ref{tab:answer_prompting_ablation} reports the results of an ablation study examining the effect of different prompting strategies in the final answer generation stage,
where the retrieved context is provided as input.
We compare three prompting strategies:
naive prompting \citep{NEURIPS2020_1457c0d6},
Chain-of-Thought (CoT) prompting \citep{NEURIPS2022_9d560961},
and CoT self-consistency \citep{DBLP:conf/iclr/0002WSLCNCZ23}.
Across all settings, the retrieval and selection results
(i.e., the input context provided to the QA model)
are kept identical, and only the answer generation prompt is varied.

The results show that the naive prompting strategy achieves the highest performance in both Token F-1 and LF-Eval.
In contrast, applying CoT leads to a consistent degradation in both metrics.
While CoT self-consistency ($n=5$) partially recovers performance compared to CoT, it still underperforms the naive prompting setting.

These trends can be attributed to the characteristics of legal QA and to the constraints imposed by our QA prompting design,
which enforces answers to be strictly grounded in the provided statutory context.
First, answers in KoBLEX are often expressed as concise legal conclusions. When CoT prompting increases the length of generated outputs, token-level overlap with the gold answer may decrease, leading to lower Token F-1 scores.
Second, during the explicit narration of reasoning steps,
CoT may encourage the model to rely on its general legal knowledge or commonsense reasoning rather than directly reflecting the provided statutory text, which can be detrimental under context-grounded constraints and negatively affect legal faithfulness as measured by LF-Eval. 
Third, although self-consistency aggregates multiple reasoning paths via majority voting, when the retrieved context is already issue-aligned and unambiguous, additional CoT sampling may introduce unnecessary variance and weaken evidence consistency.

Overall, these findings suggest that when the retrieval pipeline of \methodName{} provides precise, issue-level statutory evidence, concise, evidence-grounded answer generation via naive prompting can be more effective than reasoning-oriented prompting strategies such as CoT, both in terms of answer accuracy and legal faithfulness.

\section{Case Study}
\label{appendix:case_study}

This section presents a qualitative analysis of representative success and failure cases to provide an intuitive understanding of the behavior and limitations of \methodName{}.
The cases are categorized into three types according to the success or failure of \textit{question decomposition} and \textit{statutory context retrieval}. All examples presented in the tables are originally written in Korean, and those not provided in English in the original dataset have been machine-translated for clarity and presentation.

\paragraph{(1) Successful Case: Correct Decomposition and Retrieval.}
Table~\ref{tab:case_study_succeed} illustrates a case in which both question decomposition
and context retrieval are successfully performed.
\methodName{} accurately decomposes a complex legal question into a set of independent
atomic sub-questions and generates appropriate parametric queries for each sub-question,
thereby identifying the core statutory provisions required for answering the question
in a step-by-step manner.
Because each sub-question is explicitly paired with its corresponding statute,
the multi-hop legal reasoning process becomes transparent,
enabling issue-level verification of the reasoning steps.

\paragraph{(2) Error Analysis: Failure due to Question Decomposition.}
Table~\ref{tab:case_study_failure_decomp} presents a failure case caused by an error in the
question decomposition stage.
The initial sub-question fails to accurately capture the core legal issue of the original question,
which subsequently leads to poorly aligned parametric queries.
As a result, errors propagate through the retrieval stage,
ultimately degrading the quality of the final answer.
This case highlights that the performance of \methodName{} is highly dependent on the accuracy
of question decomposition, and that errors at this stage can have cascading effects on downstream
retrieval and answer generation.

\paragraph{(3) Error Analysis: Correct Decomposition but Erroneous Retrieval.}
Table~\ref{tab:case_study_failure_retrieve} shows a case where question decomposition is successfully performed, but an inappropriate statute is retrieved for a particular sub-question.
Although the sub-question itself is clearly defined, the generated parametric queries and retrieval results fail to align precisely with the most relevant statutory provision.
Consequently, incorrect legal evidence is accumulated during the reasoning process. This case demonstrates that, in legal question answering, question decomposition alone is insufficient, and that accurate issue-level statute alignment and selection remain critical challenges.

Overall, these case studies demonstrate that when \methodName{} operates as intended, it can decompose complex legal questions into issue-level sub-questions and systematically accumulate the corresponding core statutes, enabling interpretable and evidence-grounded legal reasoning.
At the same time, the analysis clearly reveals how errors in either the decomposition or retrieval stage can affect the entire reasoning pipeline, highlighting the need for improved decomposition robustness and more precise issue-level retrieval in future work.

\begin{table*}[ht]
\resizebox{\linewidth}{!}{%
\begin{tabular}{lp{0.85\linewidth}}
\toprule
\multicolumn{2}{l}{\textbf{Gold}} \\ \midrule
\textbf{Background} & \begin{tabular}[c]{@{}l@{}}Person A has filed a petition with the Constitutional Court against Person B's political \\ party, alleging actions in violation of the national constitution. \\ The Constitutional Court has accepted the case, and Person A is curious about \\ the timeline for the final decision regarding Person B's party and the consequences \\if a dissolution order is issued.\end{tabular} \\ \hline
\textbf{Question} & \begin{tabular}[c]{@{}l@{}}When the Constitutional Court accepts a case regarding the constitutionality of \\Person B's political party, within how many days must the final decision be rendered, \\ and what happens to Person B's party if a decision ordering its dissolution is issued?\end{tabular} \\ \hline
\textbf{Gold\_context} & \begin{tabular}[c]{@{}l@{}}{[}1{]} CONSTITUTIONAL COURT ACT / Article. 38 / Time Limit of Adjudication\\ Article 38 (Time Limit of Adjudication) The Constitutional Court shall pronounce \\the final decision within 180 days after it receives the case for adjudication: \\ Provided, That if the attendance of seven Justices is impossible due to vacancies of \\Justices, the period of vacancy shall not be counted in the period of adjudication.\\ {[}2{]} CONSTITUTIONAL COURT ACT / Article. 59 / Effect of Decision\\ Article 59 (Effect of Decision) When a decision ordering the dissolution of a political \\party is pronounced, the political party shall be dissolved.\end{tabular} \\ \midrule
\multicolumn{2}{l}{\textbf{Model Response}} \\ \midrule
\textbf{Sub-Question 1} & When the Constitutional Court accepts a case concerning the constitutionality of Party B, within how many days must the final decision be rendered? \\ \hline
\textbf{Parametric Query} & \begin{tabular}[c]{@{}l@{}}{[}1{]} Constitutional Court Act (Conclusion of Adjudication and Decision) — The \\Constitutional Court shall pronounce a final decision within 180 days from the date it \\receives the petition.\\ {[}2{]} Constitutional Court Act (Effect of a Decision Ordering Party Dissolution) — A \\decision ordering the dissolution of a political party results in the party’s assets and \\facilities being vested in the State and the prohibition of the party’s activities.\\ {[}3{]} Constitutional Court unconstitutionality review, political party dissolution, decision \\period, final decision, adjudication procedure, constitutional complaint\\ {[}4{]} Constitutional Court Act, unconstitutional political party, dissolution adjudication\\ {[}5{]} Constitutional Court Act, time limit for pronouncing a final decision\\ {[}6{]} Constitutional Court Act, political party dissolution procedure\end{tabular} \\ \hline
\textbf{Retrieved Context} & \begin{tabular}[c]{@{}l@{}}CONSTITUTIONAL COURT ACT / Article. 38 / Time Limit of Adjudication\\ Article 38 (Time Limit of Adjudication) The Constitutional Court shall (...)\end{tabular} \\ \hline
\textbf{Sub-Question 2} & If a decision ordering the dissolution of Party B is pronounced, what happens to Party B? \\ \hline
\textbf{Parametric Query} & \begin{tabular}[c]{@{}l@{}}{[}1{]} Constitutional Court Act (Effect of a Political Party Dissolution Decision) — A \\decision ordering the dissolution of a political party has the effect of terminating the \\existence and activities of the party concerned.\\ {[}2{]} Constitutional Court Act (Pronouncement of a Political Party Dissolution Decision) \\— When the Constitutional Court decides to dissolve a political party in a party \\dissolution adjudication, it shall clearly state the reasons for the decision.\\ {[}3{]} Constitutional Court Act (Filing of Petition and Adjudication Procedure) — The \\Constitutional Court shall conclude oral arguments and pronounce its decision within \\180 days from the date it receives the petition for adjudication.\\ {[}4{]} Effects of a political party dissolution decision, Constitutional Court, party \\dissolution procedure, handling of property rights, obligations, disadvantages of a \\dissolution order, starting point of legal effects\\ {[}5{]} Constitutional Court Act, effect of a political party dissolution decision\\ {[}6{]} Constitutional Court Act, legal consequences following political party dissolution\\ {[}7{]} Political Parties Act, party dissolution decision procedure\end{tabular} \\ \hline
\textbf{Retrieved Context} & \begin{tabular}[c]{@{}l@{}}CONSTITUTIONAL COURT ACT / Article. 59 / Effect of Decision\\ Article 59 (Effect of Decision) When a decision ordering the dissolution of a (...)\end{tabular} \\ \bottomrule
\end{tabular}%
}
\caption{Successful case study of \methodName{} on a multi-hop statute-based legal question.
For brevity, repeated statutory text in the retrieved context that overlaps with the gold context is abbreviated with ellipses.}
\label{tab:case_study_succeed}
\end{table*}
\begin{table*}[ht]
\resizebox{\linewidth}{!}{%
\begin{tabular}{lp{0.85\linewidth}}
\toprule
\multicolumn{2}{l}{\textbf{Gold}} \\ \midrule
\textbf{Background} & Five years ago, Person A lent money to Person B, but Person B has not repaid the debt. Concerned that the statute of limitations on the debt is approaching, Person A wishes to take legal action to interrupt the prescription period. Person A has decided to first apply for a provisional attachment. \\ \hline
\textbf{Question} & What actions can Person A take to interrupt the statute of limitations on the debt owed by Person B? \\ \hline
\textbf{Gold\_context} & \begin{tabular}[c]{@{}l@{}}{[}1{]} CIVIL ACT / Article. 168 / Causes Interrupting Extinctive Prescription\\ Article 168 (Causes Interrupting Extinctive Prescription) Extinctive prescription shall be \\interrupted in any of the following cases:1. Demand;2. (...)\\ {[}2{]} CIVIL EXECUTION ACT / Article. 279 / Request for Provisional Seizure / Para. 1\\ The following matters shall be entered in a request for provisional seizure:1. Indication \\of a requested claim, and when such requested claim is not a specific sum of money, the \\amount converted into the money;\\ 2. Indication of the facts which are to become the reasons for a provisional seizure \\under Article 277.\end{tabular} \\ \midrule
\multicolumn{2}{l}{\textbf{Model Response}} \\ \midrule
\textbf{Sub-Question 1} & Is it possible for Person A to apply for a provisional attachment in order to interrupt the statute of limitations on the debt owed by Person B? \\ \hline
\textbf{Parametric Query} & \begin{tabular}[c]{@{}l@{}}{[}1{]} Civil Act (Interruption of Extinctive Prescription) — Extinctive prescription is \\interrupted upon the occurrence of any of the following grounds: transfer or inheritance \\of a claim; (...)\\ {[}2{]} Debt, provisional attachment, interruption of extinctive prescription, creditor, debtor,\\ Civil Execution Act\\ {[}3{]} Civil Procedure Act, provisional attachment, interruption of extinctive \\prescription\\ {[}4{]} Civil Act, claim, extinctive prescription, provisional attachment\\ {[}5{]} Debtor Rehabilitation and Bankruptcy Act, effect of provisional attachment\end{tabular} \\ \hline
\textbf{Retrieved Context} & \begin{tabular}[c]{@{}l@{}}Civil Act / Article 168(2) (Causes Interrupting Extinctive Prescription)\\ Attachment, provisional attachment, or provisional disposition shall interrupt extinctive \\prescription.\end{tabular} \\ \hline
\textbf{Sub-Question 2} & Apart from applying for a provisional attachment, what other measures can Person A take to interrupt the statute of limitations on the debt owed by Person B? \\ \hline
\textbf{Parametric Query} & \begin{tabular}[c]{@{}l@{}}{[}1{]} Civil Act (Interruption of Extinctive Prescription) — Extinctive prescription shall be \\interrupted in any of the following cases: demand; attachment, provisional attachment,\\ or provisional disposition; acknowledgment of the obligation by the debtor.\\ {[}2{]} Civil Act (Demand) — Where a creditor makes a demand against the debtor, \\extinctive prescription shall be interrupted; however, if the creditor fails to take\\ further action within six months after becoming aware that the demand is groundless, \\the prescription shall resume running.\\ {[}3{]} Civil Act (Acknowledgment of Obligation) — Where the debtor acknowledges the \\existence of the obligation or expresses an intention to perform it, extinctive\\prescription shall be interrupted.\\ {[}4{]} Debt, interruption of extinctive prescription, provisional attachment, payment order, \\lawsuit, creditor, debtor\\ {[}5{]} Civil Procedure Act, interruption of extinctive prescription\\ {[}6{]} Civil Procedure Act, claim, provisional attachment\\ {[}7{]} Civil Act, claim, interruption of extinctive prescription\\ {[}8{]} Civil Act, non-performance of obligation\end{tabular} \\ \hline
\textbf{Retrieved Context} & \begin{tabular}[c]{@{}l@{}}Civil Act / Article 168(1) (Causes Interrupting Extinctive Prescription)\\ A demand shall interrupt extinctive prescription.\end{tabular} \\ \bottomrule
\end{tabular}%
}
\caption{Failure case of \methodName{} caused by error propagation from an incorrect question decomposition to subsequent retrieval steps.}
\label{tab:case_study_failure_decomp}
\end{table*}
\begin{table*}[ht]
\resizebox{\linewidth}{!}{%
\begin{tabular}{lp{0.85\linewidth}}
\toprule
\multicolumn{2}{l}{\textbf{Gold}} \\ \midrule
\textbf{Background} & Person A and Person B entered into a contract for the sale of goods. Person A agreed to supply a specific product to Person B within one month. However, Person A failed to deliver the goods by the contractual deadline. As a result, Person B intends to rescind the contract. The contract included a clause stating, In the event of a delay in performance, Person B does not require additional written notice to rescind the contract. \\ \hline
\textbf{Question} & Under what conditions can Person B rescind the contract if Person A fails to fulfill their contractual obligations, and how does the presence of other clauses in the contract affect this? \\ \hline
\textbf{Gold\_context} & \begin{tabular}[c]{@{}l@{}}{[}1{]} CIVIL ACT / Article. 544 / Delay of Performance and Rescission\\ Article 544 (Delay of Performance and Rescission) If one of the parties does not perform \\his obligation, the other party may fix a reasonable period and give peremptory notice \\demanding its performance, and may rescind the contract, if no performance is effected \\within such period: Provided, That if the obligor declares in advance his/her intention \\that he will not effect such performance, no peremptory notice shall be required. \\ {[}2{]} CIVIL ACT / Article. 105 / Optional Provisions\\ Article 105 (Optional Provisions) If the parties to a juristic act have declared an intention \\which differs from any provisions of statutes, which are not concerned with good morals \\or other social order, such intention shall prevail.\end{tabular} \\ \midrule
\multicolumn{2}{l}{\textbf{Model Response}} \\ \midrule
\textbf{Sub-Question 1} & Under what conditions can Person B rescind the contract if Person A fails to perform their contractual obligations? \\ \hline
\textbf{Parametric Query} & \begin{tabular}[c]{@{}l@{}}{[}1{]} Civil Act (Rescission of Contract) — Where the obligor delays performance of the \\obligation, the obligee may fix a reasonable period and urge performance; (...)\\ {[}2{]} Civil Act (Rescission of Contract) — Where there is a special agreement (...)\\ {[}3{]} Sale contract, delay of performance, rescission, conditions for exercising (...)\\ {[}4{]} Civil Act, rescission for delay\\ {[}5{]} Civil Act, non-performance, rescission\\ {[}6{]} Civil Act, delay of performance, right of rescission\\ {[}7{]} Commercial Act, rescission of sale contract\end{tabular} \\ \hline
\textbf{Retrieved Context} & \begin{tabular}[c]{@{}l@{}}CIVIL ACT / Article 545 (Fixed-Time Acts and Rescission)\\ Article 545 provides that where, due to the nature of the contract or the parties’ declared \\intentions, the purpose of the contract cannot be achieved unless performance is rendered \\by a specified date or within a specified period, and one party fails to perform at (...).\end{tabular} \\ \hline
\textbf{Sub-Question 2} & How does the situation change if the contract contains other clauses related to rescission? \\ \hline
\textbf{Parametric Query} & \begin{tabular}[c]{@{}l@{}}{[}1{]} Civil Act (Rescission of Contract) — Where the obligor fails to make a genuine (...)\\ {[}2{]} Civil Act (Exercise of the Right of Rescission) — The right of rescission shall (...)\\ {[}3{]} Civil Act (Principle of Priority of Special Agreements) — Where the contracting (...)\\ {[}4{]} Contract rescission, delay of performance, conflict of contractual clauses, (...)\\ {[}5{]} Civil Act, rescission for delay of performance\\ {[}6{]} Civil Act, special agreement clauses on rescission\\ {[}7{]} Civil Act, declaration of intention for rescission\\ {[}8{]} Civil Act, non-performance, right of rescission\end{tabular} \\ \hline
\textbf{Retrieved Context} & \begin{tabular}[c]{@{}l@{}}ACT ON THE REGULATION OF TERMS AND CONDITIONS / Article 5 \\ Article 5 governs the interpretation of standardized contract terms and provides guiding \\principles for resolving ambiguities or conflicts in contractual clauses.\end{tabular} \\ \bottomrule
\end{tabular} %
}
\caption{Failure case of \methodName{} where correct question decomposition is followed by erroneous statute retrieval.
Lengthy parametric queries and retrieved contexts are partially omitted and indicated with (...).}
\label{tab:case_study_failure_retrieve}
\end{table*}

\end{document}